%% file: root.tex
\documentclass[letterpaper, 10 pt, conference]{ieeeconf}  

\IEEEoverridecommandlockouts                              

\usepackage{graphicx} 
\usepackage{amsmath} 
\usepackage{amssymb}  
\usepackage{siunitx}
\usepackage{multirow}
\usepackage{booktabs}
\usepackage{url}

\title{\LARGE \bf
Thegra: Graph-based SLAM  for Thermal Imagery
}

\author{Anastasiia Kornilova$^{1*}$, Ivan Moskalenko$^{1*}$, Arabella Gromova$^{1}$, 
\\ Gonzalo Ferrer$^{1}$ and Alexander Menshchikov$^{1}$%
\thanks{$^{1}$Skolkovo Institute of Science and Technology, Moscow, Russia.
        {\tt\small \{A.Kornilova, I.Moskalenko, A.Gromova, G.Ferrer, A.Menshchikov\}@skoltech.ru}}%
\thanks{* Indicates equal contribution.}%
}

\begin{document}

\maketitle
\thispagestyle{empty}
\pagestyle{empty}

\input{src/00_abstract}

\input{src/01_introduction}
\input{src/02_related}
\input{src/03_method}
\input{src/05_experiments}
\input{src/09_conclusion}

\bibliographystyle{ieeetr}
\bibliography{references}

\end{document}

%% file: src/00_abstract.tex
\begin{abstract}

Thermal imaging provides a practical sensing modality for visual SLAM in visually degraded environments such as low illumination, smoke, or adverse weather. However, thermal imagery often exhibits low texture, low contrast, and high noise, complicating feature-based SLAM. In this work, we propose a sparse monocular graph-based SLAM system for thermal imagery that leverages general-purpose learned features~--- the SuperPoint detector and LightGlue matcher, trained on large-scale visible-spectrum data to improve cross-domain generalization. To adapt these components to thermal data, we introduce a preprocessing pipeline to enhance input suitability and modify core SLAM modules to handle sparse and outlier-prone feature matches. We further incorporate keypoint confidence scores from SuperPoint into a confidence-weighted factor graph to improve estimation robustness. Evaluations on public thermal datasets demonstrate that the proposed system achieves reliable performance without requiring dataset-specific training or fine-tuning a desired feature detector, given the scarcity of quality thermal data. Code will be made available upon publication.

\end{abstract}

%% file: src/01_introduction.tex
\section{Introduction}

Thermal imaging has become a practical sensing modality for visual odometry and SLAM (Simultaneous Localization and Mapping), particularly in environments with limited illumination, visual obscurants such as smoke or dust, or adverse weather conditions. Nonetheless, the specific characteristics of thermal imagery~--- such as low texture, limited contrast, non-uniformity correction (NUC) artifacts, and sensor noise,~--- pose challenges to conventional visual SLAM systems, which are typically designed for photometrically rich RGB input.

To address these challenges, prior studies have proposed a variety of approaches. Multimodal systems combining thermal cameras with visible-light sensors~\cite{poujol2016visible, khattak2019visual, qin2023bvt}, LiDAR~\cite{shin2019sparse, de20223d}, or inertial measurement units (IMUs)~\cite{papachristos2018thermal, delaune2019thermal} have demonstrated increased robustness, while adaptations of filtering-based and dense visual odometry techniques have allowed operation on raw thermal data~\cite{papachristos2018thermal}. The present paper focuses on thermal-only data, revisiting the visual-SLAM pipeline highlighting the blocks that require updating or modifications.

In more recent work, pose graph SLAM frameworks~\cite{jiang2022thermal} and learning-based keypoint detectors have been applied to improve tracking stability and map reliability~\cite{keil2024towards}. However, these approaches remain constrained by both {\em sensor-specific} dependencies and {\em limited training data} compared to large-scale visible-spectrum datasets. Learning-based methods in particular may suffer a performance drop upon domain transfer between different thermal cameras or environments, as observed in experimental evaluations, ruling out any custom thermal-data-driven approach to test.

\begin{figure}[!htb]
    \centering
    \includegraphics[width=0.48\textwidth]{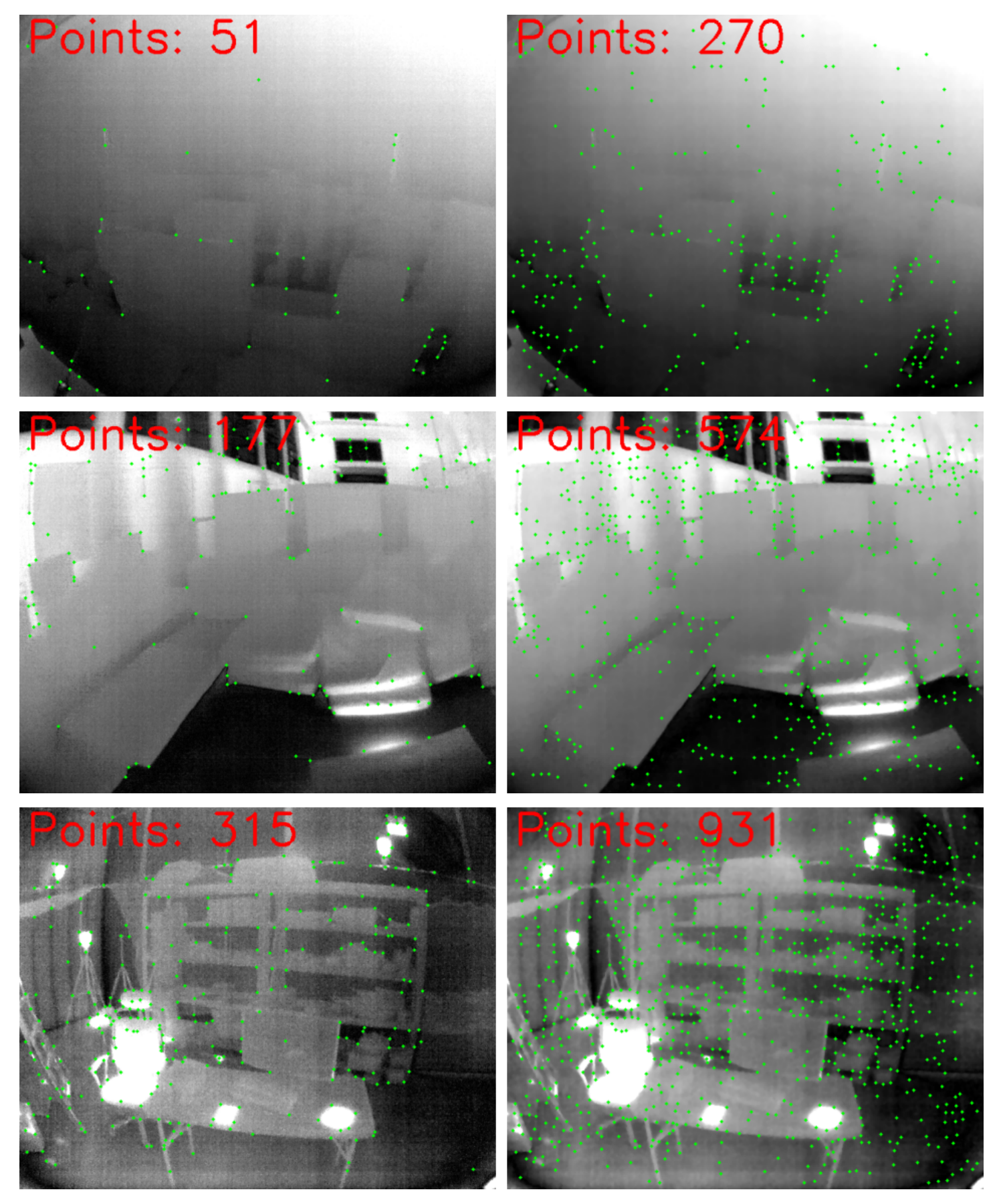} 
    \caption{Visualization of SuperPoint feature detections. Left: Detections on images subjected to basic preprocessing (normalization only). Right: Detections on images processed using the proposed preprocessing technique, which includes Chambolle denoising, histogram equalization, and median filtering.}
    \label{fig:sp_detections}
\end{figure}

This paper presents a sparse monocular SLAM system designed for thermal imagery, leveraging general-purpose learned features for improved cross-domain generalization, where training data is scarce and the variability depending on the sensor used is considerable. Unlike dense or end-to-end approaches for thermal SLAM, our method builds upon the widely used SuperPoint detector~\cite{detone2018superpoint} and LightGlue matcher~\cite{lindenberger2023lightglue}, models trained on large-scale visible-spectrum data, to ensure robustness across diverse thermal sensors and environments. To bridge the domain gap between thermal and visual imagery, we investigate different preprocessing techniques and introduce a preprocessing pipeline that enhances input suitability for these off-the-shelf networks.
Finally, we incorporate a weighted factor graph optimization that utilizes keypoint confidence scores from SuperPoint, that increases tracking stability. Experimental evaluation on multiple thermal datasets demonstrates that our system achieves consistent performance and robust tracking without requiring dataset-specific retraining or fine-tuning, further demonstrating its ability to generalize across diverse thermal environments, unlike some comparative methods which exhibit limitations in stability or scale estimation.

The main contributions of this work are:
\begin{itemize}
\item A thermal sensor-invariant SLAM system that utilizes general-purpose learned SuperPoint detector and LightGlue matcher for improved generalization across diverse thermal sensors and environments;
\item Evaluation of preprocessing techniques for enhancing feature detection and matching in thermal imagery;
\item A confidence-weighted factor graph optimization method based on detector confidence scores that improves tracking stability and overall system robustness.
\end{itemize}

%% file: src/02_related.tex
\section{Related Work}
Thermal imaging introduces significant challenges for Visual Inertial Odometry (VIO) and SLAM systems, such as low contrast, limited texture, high levels of sensor noise, and artifacts produced by NUC processes, including stripe noise. To address these issues, many systems incorporate additional sensing modalities that enhance robustness and enable metric scale recovery.

Multimodal sensor configurations are frequently adopted to compensate for the deficiencies of thermal data. Stereo thermal setups~\cite{mouats2015thermal}, visible-thermal stereo pairs~\cite{poujol2016visible, khattak2019visual, qin2023bvt}, and fusion with LiDAR~\cite{shin2019sparse, de20223d} or RADAR~\cite{zhang20234drt} provide complementary spatial or spectral information, improving feature detectability and depth estimation. Additionally, the integration of inertial measurement units (IMUs)~\cite{papachristos2018thermal, delaune2019thermal} enables compact and reliable thermal-inertial SLAM systems, particularly beneficial for lightweight aerial platforms. Unlike LiDAR and RADAR, IMUs are immune to environmental degradations such as dust or smoke, though they require careful drift compensation through advanced filtering techniques.

This review surveys recent progress in thermal SLAM and VIO, focusing on three key areas: thermal image preprocessing, graph-based SLAM frameworks, and alternative SLAM/VIO methods.

\subsection{Thermal Preprocessing for Localization}
Thermal image preprocessing is essential in SLAM and VIO systems to improve feature quality and density. Common techniques include contrast enhancement, denoising, normalization, and stripe noise removal, addressing issues like NUC artifacts and low dynamic range. However, few studies directly assess the impact of these techniques on SLAM/VIO performance.

In recent literature, various preprocessing strategies have been employed. For instance, the work by Lv et al.~\cite{lv2024visual} applies a bilateral filter for image denoising, a bandpass filter for stripe noise suppression, and CLAHE for enhancing image contrast. Similarly, CLAHE is utilized in the approach proposed by Keil et al.~\cite{keil2024towards}. In contrast, Chen et al.~\cite{chen2022eil} adopt standard histogram equalization to achieve contrast enhancement. A distinct methodology is presented by Jiang et al.~\cite{jiang2022thermal}, who employ Singular Value Decomposition (SVD)-based processing for the suppression of stripe noise artifacts. Additionally, some approaches convert thermal images into binarized edge representations using a Difference of Gaussians (DoG) filter, as demonstrated in the work by Wang et al.~\cite{wang2023edge}.

Given the diversity of preprocessing techniques and their potential impact on feature extraction and tracking, further investigation into their comparative effectiveness remains an open area of research within the domain of thermal SLAM/VIO.

\subsection{Graph-Based Methods}
Graph-based SLAM formulates the localization and mapping problem as an optimization task over a sparse factor graph, where nodes represent robot poses and landmarks, and edges encode spatial constraints derived from sensor measurements. This structure facilitates efficient optimization with global consistency and supports the integration of diverse sensing modalities, making it suitable for large-scale and long-term applications.

ORB-SLAM~\cite{mur2015orb} and its extension ORB-SLAM3~\cite{campos2021orb} are canonical examples of graph-based SLAM systems. ORB-SLAM3 incorporates monocular, stereo, RGB-D, and visual-inertial inputs, and supports real-time tracking, loop closure, and relocalization. It relies on ORB~\cite{rublee2011orb} features for matching and achieves global consistency through pose graph optimization and bundle adjustment.

Several adaptations of graph-based SLAM have been proposed for thermal imaging. FirebotSLAM~\cite{van2023firebotslam} extends ORB-SLAM3 to thermal data by dynamically scaling 16-bit radiometric images to 8-bit to enhance feature visibility. It employs SURF~\cite{bay2008speeded} features with extended BRIEF~\cite{calonder2010brief} descriptors and integrates all SLAM modules within the ORB-SLAM3 framework.

DVT-SLAM~\cite{wang2021dvt} introduces a data fusion approach based on DVT-GAN, which leverages contrastive learning to align thermal and visible domains. The fused images are converted to 8-bit and processed using ORB features in a graph-based SLAM pipeline.

Saputra et al.~\cite{saputra2021graph} propose a thermal-inertial SLAM system that utilizes Mixture Density Networks (MDNs) for probabilistic pose estimation and loop closure. The system incorporates hallucinated RGB images and learned global descriptors for place recognition.

Keil et al.~\cite{keil2024towards} design a thermal SLAM system based on the Multi-Camera SLAM (MCSLAM) framework. They replace ORB features with SuperPoint~\cite{detone2018superpoint} descriptors and use Gluestick~\cite{pautrat2023gluestick}, a learned matching module, to improve robustness across thermal conditions. Loop closure is supported by a BoW vocabulary trained on Gluestick-enhanced SuperPoint features.

In~\cite{jiang2022thermal}, a thermal-inertial SLAM system fuses thermal imagery with inertial data within a graph-based framework. Keyframes are connected via visual-inertial constraints and optimized through pose graph methods. ThermalRAFT, a deep optical flow network, is used for feature tracking.

Shin et al.~\cite{shin2019sparse} introduce a sparse thermal-infrared SLAM method that fuses sparse LiDAR and thermal image features for 6-DOF motion estimation. Tracking is based on thermographic error models, and pose refinement is conducted via keyframe-based optimization. Loop closure is achieved using a Bag-of-Words approach followed by global pose graph optimization.

To date, research on graph-based SLAM in the thermal imaging domain remains limited, with most existing implementations relying on hand-crafted features, thus highlighting the potential for further methodological advancements.

\subsection{Other Methods}
In addition to graph-based approaches, various alternative methods for SLAM and VIO have been developed. Filtering-based techniques, such as the Extended Kalman Filter (EKF), employ recursive state estimation and are effective for real-time applications. A prominent example is ROVIO~\cite{bloesch2015robust}, which integrates inertial and visual data within an EKF framework using direct photometric alignment of image patches. Another notable approach is the on-manifold EKF-based estimator OpenVINS~\cite{geneva2020openvins}, which features a modular and extensible architecture supporting diverse stereo and multi-camera configurations.

Methods utilizing sliding window optimization also merit special attention. Popular methods in this category include DSO~\cite{engel2017direct} and SVO~\cite{forster2014svo}. DSO operates via photometric bundle adjustment on selected regions, while SVO combines sparse initialization with direct tracking. Direct or semi-direct methods, such as SVO and DSO, optimize photometric consistency directly on pixel intensities, often achieving robustness in low-texture environments, although at the cost of increased computational complexity.

Several methods specifically target thermal imaging scenarios. Papachristos et al.~\cite{papachristos2018thermal} adapted ROVIO for thermal data by fusing long-wave infrared imagery with inertial cues, achieving real-time tracking in visually degraded environments like smoke-filled rooms. Zhao et al.~\cite{zhao2020tp} proposed the ThermalPoint (TPTIO) system, which employs deep feature detection and KLT tracking. Learning-based enhancements, such as Self-TIO~\cite{lee2024self}, employ a hybrid tracker that merges self-supervised learning with KLT tracking. DeepTIO~\cite{saputra2020deeptio} uses a hallucination network to synthesize visual-like features from thermal input, improving feature diversity and leveraging selective fusion across multiple modalities. Khattak et al.~\cite{khattak2020keyframe} minimize error on raw 14-bit thermal images using pyramidal representations, while Lv et al.~\cite{lv2024visual} enhance thermal imagery through adaptive filtering and extract both point and line LBD descriptors~\cite{zhang2013efficient}.

Thus, while alternative methods offer advantages in real-time performance and computational efficiency, they often lack the global consistency and robustness provided by graph-based approaches, particularly in large-scale and long-term scenarios.

%% file: src/03_method.tex
\section{Methodology}

This work presents a sparse monocular graph-based SLAM system designed for thermal imagery, addressing the particular challenges of low-texture thermal data through several key adaptations. The system employs state-of-the-art general-purpose neural keypoint and neural matcher, augmented with a specialized preprocessing pipeline to improve feature detection and matching performance (Section~\ref{sec:preprocessing}). Since standard visual SLAM does not generalize well as demonstrated in experiments the proposed architecture incorporates modifications across components to handle sparse feature sets and outlier-prone matches characteristic of thermal imagery, architecture overview and modifications are presented in Section~\ref{sec:architecture}. Furthermore, system robustness is enhanced through a weighted factor graph formulation that utilizes keypoint confidence scores from SuperPoint (Section~\ref{sec:weighted-fgo}). 

\subsection{Keypoint Detection and Matching}
\label{sec:det_match}

\begin{figure}
    \centering
    \includegraphics[width=0.44\textwidth]{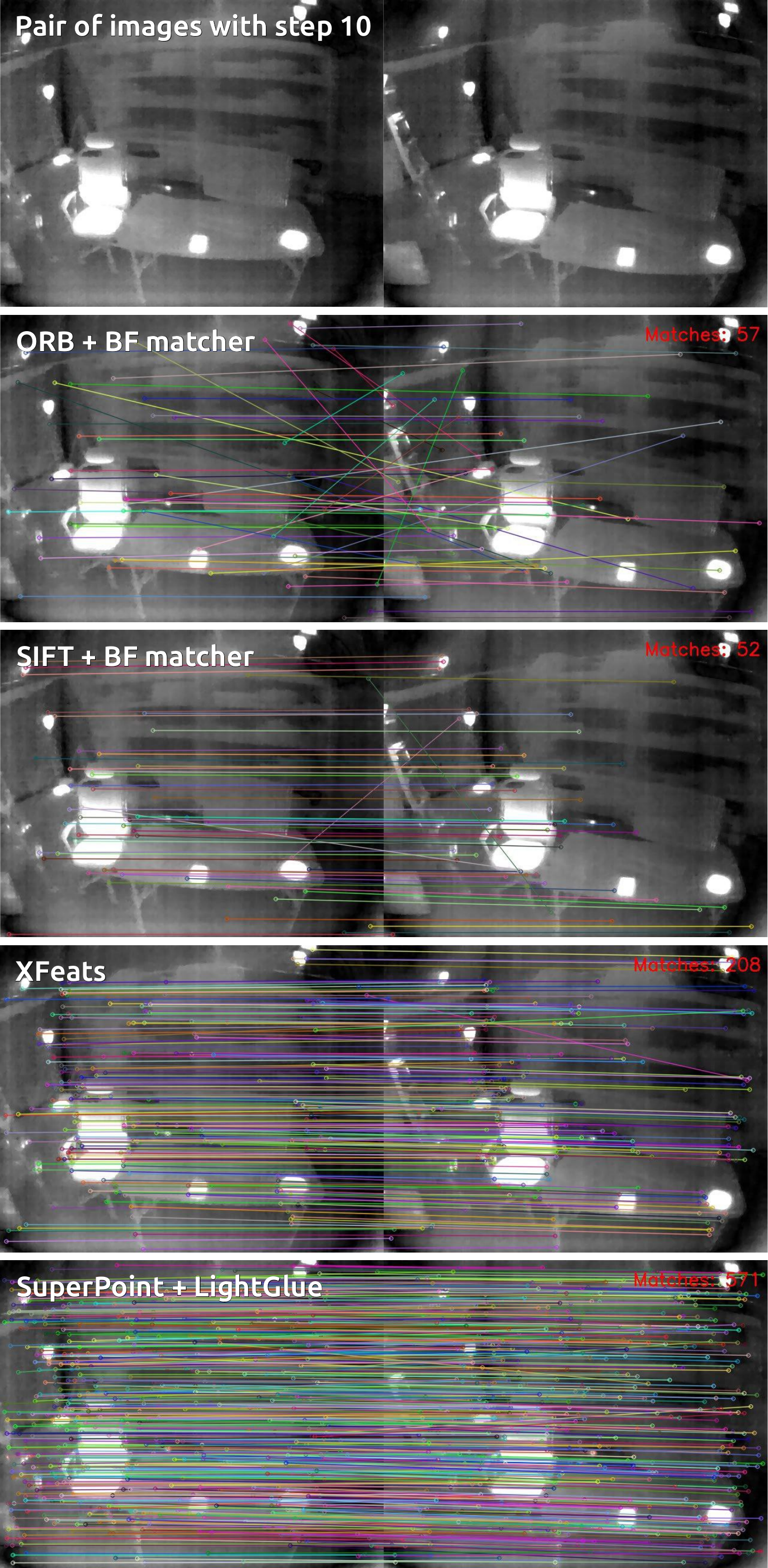}
    \caption{Comparison of combination of detectors and matchers for a pair of thermal images, both handcrafted and learnable approaches.}
    \label{fig:det_match_comp}
\end{figure}

Our architecture centers on the pre-trained SuperPoint keypoint detector and LightGlue matcher due to their robustness and generalization capabilities. These general-purpose models benefit from large-scale training on diverse visible-spectrum datasets, whereas thermal-specific alternatives are typically limited to smaller, often sensor-specific thermal datasets. This enables robust performance across diverse thermal domains and camera models, mitigating the constraints of thermal data scarcity and sensor-specific bias. Furthermore, as illustrated in Fig.~\ref{fig:det_match_comp}, both popular handcrafted methods (e.g., ORB, SIFT, BF matcher) and other modern learnable approaches (e.g., XFeat~\cite{potje2024xfeat}) were found to be less effective in our preliminary empirical evaluations, producing fewer matches and less stable correspondences.

\subsection{Thermal imagery preprocessing}
\label{sec:preprocessing}

Thermal imagery presents particular challenges, including low contrast, high noise, and lack of texture, that degrade even state-of-the-art feature extraction methods. This domain gap between thermal and visible-spectrum data necessitates preprocessing to improve input suitability. We employ classical, non-learned image enhancement techniques due to their interpretability and alignment with the physically-characterized noise properties of thermal sensors. These methods improve both the quantity and quality of detectable features as demonstrated in Fig.~\ref{fig:sp_detections}.

The implemented preprocessing pipeline sequentially applies a configurable sequence of filters to each input frame. The considered filter types include:

\begin{itemize}
    \item \textbf{Histogram Equalization and CLAHE.} To improve global and local contrast, we employ classical histogram equalization~\cite{pizer1987adaptive} as well as Contrast-Limited Adaptive Histogram Equalization (CLAHE)~\cite{zuiderveld1994contrast}. The CLAHE filter partitions the image into tiles and redistributes intensities within each tile to avoid over-amplification of noise.
    
    \item \textbf{Bandpass Filtering.} In the frequency domain, a bandpass filter~\cite{gonzales1987digital} is used to suppress low-frequency gradients and high-frequency noise. The filter computes the 2D Fourier transform of the image, applies a binary mask that retains frequencies within a specified radial range, and reconstructs the filtered image via inverse transform.
    
    \item \textbf{Edge-Preserving Smoothing.} To reduce noise while preserving important edge information, we utilize both bilateral filtering~\cite{tomasi1998bilateral} and total variation (TV) denoising via the Chambolle algorithm~\cite{chambolle2004algorithm}. Bilateral filtering combines intensity and spatial closeness for edge-aware blurring, while TV denoising minimizes a global energy functional to suppress noise without oversmoothing.
    
    \item \textbf{Median Filtering.} Median filtering~\cite{huang1979fast} is included as a robust technique for removing impulse noise. It replaces each pixel value with the median of its local neighborhood, effectively eliminating outlier intensities.
\end{itemize}

One of the main contributions of the present paper is the investigation on best pre-processing techniques for the downstream SLAM task, requiring classical techniques due to the scarcity of aligned thermal data.

\subsection{SLAM architecture overview}
\label{sec:architecture}

The proposed system adopts a modular architecture analogous to the monocular pipeline of ORB-SLAM3~\cite{campos2021orb}, comprising core components for map initialization, tracking, keyframe selection, local mapping, and factor-graph optimization based on reprojection error. Processing occurs per frame: each thermal image undergoes preprocessing to improve feature discriminability as describe in Sec.~\ref{sec:preprocessing}. To mitigate challenges associated with low-textured thermal imagery, such as sparse and unstable feature correspondences, key modifications emphasize prudent keypoint selection and robust outlier rejection, implemented through both heuristic strategies and optimization-based approaches. The following subsections detail system components as well as critical modifications.

\textbf{Map initialization} To ensure robust map initialization despite a lower number of features and potential outliers in matches in thermal imagery, several key modifications were implemented. First, if the number of correspondences falls below a predefined threshold and the temporal gap exceeds a specified limit, the reference frame is incrementally advanced to ensure sufficient parallax and feature overlap. Second, the essential matrix is robustly estimated using the MAGSAC algorithm~\cite{barath2019magsac}, followed by an iterative minimal-set solver for RANSAC~\cite{rosten2010improved} to further enhance model stability. Finally, a full bundle adjustment jointly optimizes both camera poses and 3D points to produce a geometrically consistent initial map.

\textbf{Tracking} The tracking module performs initial estimation of the camera pose for each incoming frame, refines it using map information, manages keyframe insertion, and maintains map quality through points culling.

\subsubsection{Initial Pose Estimation}
For each incoming frame, the system detects keypoints and computes descriptors using SuperPoint~\cite{detone2018superpoint}. These features are then matched to the current reference keyframe using LightGlue~\cite{lindenberger2023lightglue}. The resulting 2D--2D correspondences provide the foundation for an initial pose estimate. This estimate is obtained through motion-only bundle adjustment, which optimizes the camera pose while keeping the 3D structure of the map fixed.

\subsubsection{Pose Refinement}
Following the initial pose estimation, the system performs a refinement step using 2D--3D correspondences obtained by projecting all existing map points into the current frame. Each projection undergoes a robust visibility evaluation to determine its eligibility for matching. Specifically, a point is considered visible if: (i) its projection lies within the image boundaries; (ii) the angle between the current viewing direction and the point’s average observation direction is less than \ang{60}; (iii) its depth falls within a predefined, scale-invariant visibility range. If a point passes these checks, it is matched to nearby unmatched SuperPoint~\cite{detone2018superpoint} features based on descriptor similarity. The resulting set of 2D--3D matches is used in a second bundle adjustment step to refine the pose estimate with greater accuracy.

\textbf{Map Points Culling} To prevent map degradation over time, each 3D point undergoes continual evaluation based on its trackability. Specifically, a point must be successfully observed in at least 25\% of the frames in which it is predicted to be visible. Points that fail to meet this criterion are culled from the map, ensuring that only stable, well-constrained landmarks contribute to pose estimation and map growth. The system continuously evaluates whether the current frame warrants insertion as a new keyframe. Two criteria are used: (i) the cumulative rotational difference (in Euler angles) from the last keyframe exceeds \ang{15}; (ii) the normalized translation, computed as the ratio between the camera displacement since the last keyframe and the median scene depth, exceeds 0.03. When either condition is satisfied, a new keyframe is created to preserve tracking accuracy and maintain sufficient map density.

\textbf{Keyframe Insertion and Map Expansion}
When a new keyframe is introduced into the system, it is immediately incorporated into the covisibility graph as a new node. In this case, the new node is connected to such previous nodes with which the number of common observed 3D points exceeds threshold. To expand the map, feature correspondences between the new keyframe and its connected neighbors in the covisibility graph are established using LightGlue~\cite{lindenberger2023lightglue}. Prior to triangulation, correspondences that violate the epipolar constraint are discarded to ensure geometric consistency. The remaining matches are used to triangulate candidate 3D points, each of which is then validated based on several criteria. A point is retained only if it exhibits positive depth in both views, sufficient parallax between observing cameras, low reprojection error, and consistency with the expected scale.

\textbf{Local Bundle Adjustment}
To refine the local map structure and improve pose accuracy, a local bundle adjustment is performed upon the insertion of each new keyframe. This optimization jointly updates the pose of the current keyframe, its immediate neighbors in the covisibility graph, and the 3D positions of all shared map points. Additional keyframes that observe these points but are not directly connected to the current keyframe are included as fixed nodes to maintain global consistency. Outlier observations are identified and excluded both during the intermediate stages and at the conclusion of the optimization.

\subsection{Factor Graph Optimization with Weighted Scores}
\label{sec:weighted-fgo}

Backend of the developed system resembles factor-graph approach, proposed in ORB SLAM,  based on Bundle Adjustment (BA) in different modes~--- motion-only BA, local BA, full BA. It jointly optimizes the 3D positions of map points $X_{w,j} \in \mathbb{R}^3$ and the poses of keyframes $T_{iw} \in \text{SE}(3)$, such that the reprojection error with respect to the observed keypoints $x_{i,j} \in \mathbb{R}^2$ is minimized.

Key extension of this module in our work is~--- in order to account for detection noise and outliers, we minimize a robustified cost function based on squared confidence score from SuperPoint to provide insights for optimizer about feature relevance:

\begin{equation}
C = \sum_{i,j} \rho_h \left( e_{i,j}^T \Omega_{i,j} e_{i,j} \right),
\end{equation}

where $\rho_h$ is the Huber loss, and $\Omega_{i,j} = \sigma_{i,j}^2 I$ is a confidence-weighted information matrix derived from the SuperPoint detection score $\sigma_{i,j}$.

%% file: src/05_experiments.tex
\section{Experiments}

Our experimental evaluation assesses the proposed system through three primary analyses: (1) a comparative evaluation of preprocessing techniques to quantify their impact on feature detection and matching performance; (2) a benchmarking study comparing the overall SLAM system against existing visual and thermal SLAM approaches; and (3) an ablation study examining the contribution of confidence-weighted factor graph optimization using SuperPoint keypoint scores.

\subsection{Datasets}

Evaluation is conducted on two established thermal SLAM benchmarks: the ViViD++ dataset~\cite{lee2022vivid++} and the dataset proposed in ROVTIO~\cite{flemmen2021rovtio}. Both provide synchronized thermal imagery and high-precision ground truth trajectories and are widely used benchmarks for assessing thermal VIO and SLAM performance~\cite{lee2024self, chen2023thermal, jiang2022thermal, flemmen2021rovtio}. ViViD++ includes both handheld and vehicular sequences; for our experiments, we focus exclusively on the handheld sequences, which span both indoor and outdoor environments. The dataset from~\cite{flemmen2021rovtio} consists of several indoor sequences captured using a Vicon tracking system. These sequences are specifically designed to challenge SLAM algorithms by including objects with controlled variations in thermal emissivity, resulting in diverse infrared appearance profiles. The authors have made available several sequence sets—namely sync, \textit{lt1}, \textit{lt2}, \textit{lt3}, \textit{alt1}, and \textit{alt2}. For the purposes of our evaluation, we utilized temporally segmented subsequences of these datasets, each approximately 30 seconds in duration, as provided by the original authors.

\subsection{Evaluation of Thermal Imaging Preprocessing}
\label{sec:preprocessing_eval}

This study evaluates preprocessing techniques for thermal images designed to mitigate the domain gap between thermal imagery and visible-spectrum-trained models (SuperPoint and LightGlue) by increasing the number of reliable feature matches. The evaluation protocol processes image pairs, sampled at a fixed temporal interval from thermal sequences, with a candidate preprocessing filter or filter chain. Each pair undergoes feature extraction using SuperPoint followed by matching with LightGlue. The number of successful correspondences is recorded and compared across multiple filter configurations to quantify preprocessing effectiveness.

\begin{figure}[!htb]
    \centering
    \includegraphics[width=0.48\textwidth]{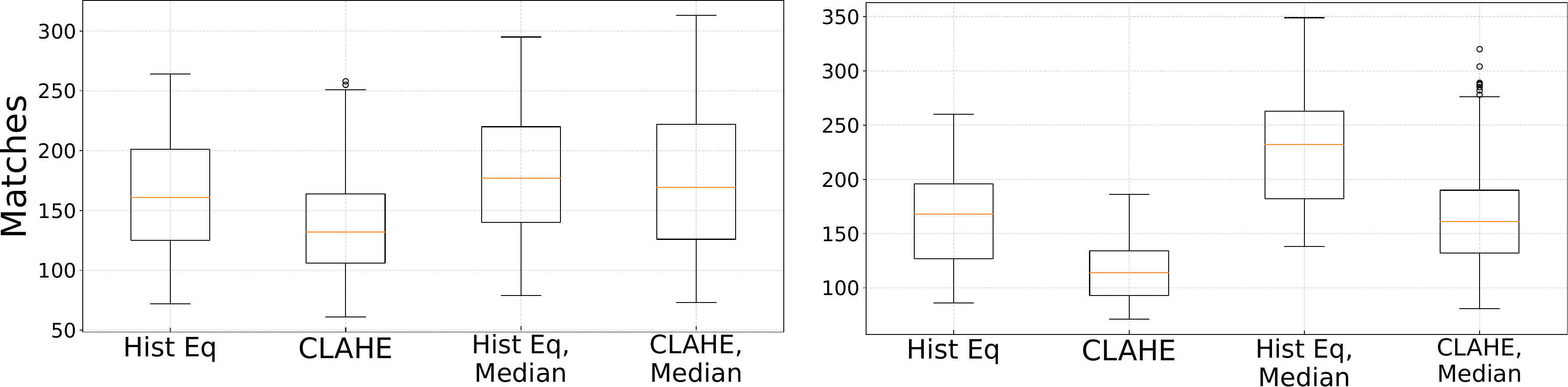} 
    \caption{Comparison of contrast enhancement techniques. Left: results on the dataset from ROVTIO. Right: results on the ViViD++ dataset.}
    \vspace{5mm}
    \label{fig:eq_clahe}
\end{figure}

In the initial phase of the evaluation, we examined contrast enhancement techniques, namely standard Histogram Equalization and Contrast Limited Adaptive Histogram Equalization (CLAHE). Each of these methods was also tested in combination with a Median filter. For Histogram Equalization, the cumulative pixel count threshold was set to 10000. In the case of CLAHE, we employed a kernel size of 8, and a clip limit of 2 was chosen to suppress noise, as values in the range of 2 to 5 are typically recommended. The Median filter was configured with a kernel size of 3. The results for the dataset from~\cite{flemmen2021rovtio} and the ViViD++ benchmark~\cite{lee2022vivid++} are presented in Fig.~\ref{fig:eq_clahe}. The findings indicate that standard Histogram Equalization slightly outperforms CLAHE in terms of feature match quantity. Additionally, the application of a Median filter subsequent to contrast enhancement yielded a significant improvement in performance.

In the second phase of our experiments, we assessed the impact of additional filtering strategies on thermal image sequences by introducing three edge-preserving or frequency-selective filters: a Bandpass filter, the Chambolle Total Variation denoising algorithm~\cite{chambolle2004algorithm}, and the Bilateral filter. These were applied in combination with the previously established baseline (Histogram Equalization followed by Median filtering) which demonstrated superior performance in the initial evaluation phase. The tested filter combinations were: Chambolle + Histogram Equalization + Median, Bilateral + Histogram Equalization + Median, and Bandpass + Histogram Equalization + Median.

\begin{figure}[!htb]
    \centering
    \includegraphics[width=0.48\textwidth]{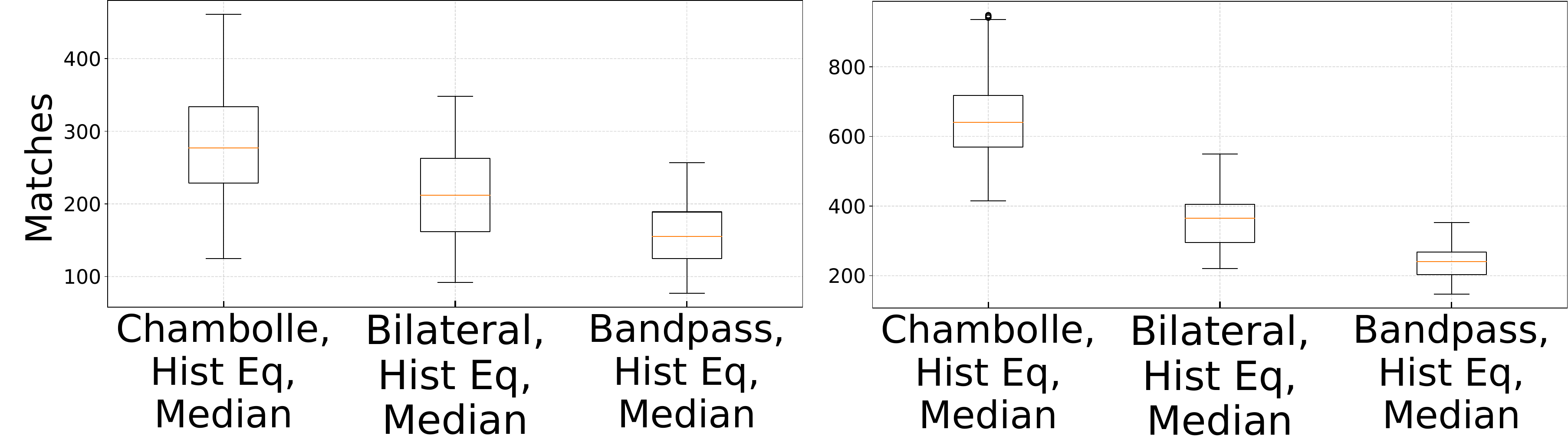} 
    \caption{Comparison of the Chambolle, Bilateral, and Bandpass filters. Left: results on the dataset from ROVTIO. Right: results on the ViViD++ benchmark.}
    \vspace{5mm}
    \label{fig:cham_bil_band}
\end{figure}

The Chambolle filter was applied with a total variation weight of 4, which was empirically chosen to provide effective noise suppression while preserving edge detail critical for structural interpretation. The Bilateral filter was configured with a spatial diameter of 4 and a spatial sigma of 35. These values were selected to ensure local smoothing while retaining edge integrity, particularly useful in suppressing small-scale thermal noise without blurring object boundaries. For the Bandpass filter, the frequency bounds were set from 0 to 87. This configuration primarily attenuates high-frequency noise components, while retaining both mid- and low-frequency information. Evaluation was conducted on the same two thermal sequences used in the first phase. The evaluation outcomes for the dataset from~\cite{flemmen2021rovtio} and the ViViD++ benchmark~\cite{lee2022vivid++} are illustrated in Fig.~\ref{fig:cham_bil_band}. The Chambolle + Histogram Equalization + Median filter bundle consistently yielded the highest number of feature matches across both sequences. As a result, this configuration was selected for use in all subsequent experiments.

\subsection{Comparison with SLAM systems}
We conducted a comparative evaluation of our method against several state-of-the-art approaches commonly employed in the field of thermal VIO and SLAM, specifically ORB-SLAM3 (a graph-based SLAM system)~\cite{campos2021orb}, ROTIO (a thermal-adapted variant of ROVIO)~\cite{flemmen2021rovtio}, and DSO (a direct method)~\cite{engel2017direct}. These algorithms are widely adopted as benchmarks in recent literature on thermal VIO/SLAM systems~\cite{lee2024self, van2023firebotslam, wang2023edge, chen2022eil, zhao2020tp}.

All baseline methods were evaluated using their publicly available implementations and default parameter settings. For ROTIO, we retained the original metric scale as derived from its visual-inertial pipeline. In contrast, for the monocular ORB-SLAM3 and DSO, trajectory scale correction was performed using the evo framework~\cite{grupp2017evo}, as was also necessary for our monocular method. To ensure a fair comparison, input thermal images were preprocessed uniformly using Chambolle filtering, histogram equalization, and median filtering for all methods, with the exception of ROTIO, which is designed to operate directly on raw thermal input.

System performance was evaluated using classical Absolute Trajectory Error (ATE). Given that certain SLAM algorithms are susceptible to tracking failures on some sequences, we also report the percentage of successfully tracked frames. For the ROVTIO dataset~\cite{flemmen2021rovtio}, which comprises multiple subsequences, evaluation was performed independently on each subsequence. Afterwards, both mean and median error metrics across all subsequences were reported. Results are summarized in Table~\ref{table:rovtio_results}. For the ViViD++~\cite{lee2022vivid++} dataset, we employed the set of sequences provided by the original authors. Detailed results for each sequence are presented in Table~\ref{table:vivid_results}.

\begin{table}[!htb]
\centering
\caption{Evaluation on ROTIO dataset}
\label{table:rovtio_results}
\large
\resizebox{\linewidth}{!}{
\begin{tabular}{|l||cc|cc|cc|cc|} 
\hline
\multirow{2}{*}{Sequence} & \multicolumn{2}{c|}{ORB SLAM3} & \multicolumn{2}{c|}{ROTIO} & \multicolumn{2}{c|}{DSO} & \multicolumn{2}{c|}{Ours} \\
\cline{2-9}
& ATE & \% & ATE & \% & ATE & \% & ATE & \% \\
\hline
alt1 & -- & 93.06 & 22.33 & 100 & -- & 92.38 & \textbf{0.14} & 100 \\
\hline
alt2 & 0.85 & 100 & 0.31 & 100 & 0.84 & 100 & \textbf{0.30} & 100 \\
\hline
sync & -- & 92.95 & \textbf{0.13} & 100 & 0.44 & 100 & 0.15 & 100 \\
\hline
lt1 & -- & 98.53 & 0.21 & 100 & 0.18 & 100 & -- & 98.75 \\
\hline
lt2 & -- & 85.12 & 3.44 & 100 & 0.51 & 100 & \textbf{0.14} & 100 \\
\hline
lt3 & -- & 87.18 & 13.40 & 100 & 0.52 & 100 & \textbf{0.13} & 100 \\
\hline
\end{tabular}
}
\end{table}

On the dataset introduced in ROVTIO~\cite{flemmen2021rovtio}, the proposed method achieves the lowest absolute trajectory error in most evaluated sequences. ORB-SLAM3 fails to complete full trajectories in the majority of cases due to tracking loss, while DSO and the method ROTIO exhibit higher trajectory instability. It is also worth noting that ROTIO often fails to accurately recover the scale of the trajectory, which is reflected in a higher mean ATE value.

\begin{table}[!htb]
\centering
\caption{Evaluation on ViViD++ dataset}
\label{table:vivid_results}
\large
\resizebox{\linewidth}{!}{
\begin{tabular}{|l||cc|cc|cc|cc|} 
\hline
\multirow{2}{*}{Sequence} & \multicolumn{2}{c|}{ORB SLAM3} & \multicolumn{2}{c|}{ROTIO} & \multicolumn{2}{c|}{DSO} & \multicolumn{2}{c|}{Ours} \\
\cline{2-9}
& ATE & \% & ATE & \% & ATE & \% & ATE & \% \\
\hline
dark\_aggresive & -- & 26.70 & 72.43 & 100 & 0.65 & 100 & \textbf{\textless 0.01} & 100 \\ 
dark\_robust & -- & 28.5 & \textbf{0.14} & 100 & 0.35 & 100 & 0.48 & 100 \\
dark\_unstable & -- & 23.9 & 28.42 & 100 & 0.75 & 100 & \textbf{0.73} & 100 \\ 
global\_aggresive & -- & 0 & 7.15 & 100 & 0.68 & 100 & \textbf{0.01} & 100 \\ 
global\_robust & -- & 0 & \textbf{0.20} & 100 & 0.66 & 100 & 0.27 & 100 \\ 
global\_unstable & -- & 0 & 1.10 & 100 & \textbf{0.63} & 100 & -- & -- \\ 
local\_aggressive & -- & 17.5 & 64.45 & 100 & 0.26 & 27.8 & \textbf{0.01} & 100 \\ 
local\_robust & 0.35 & 100 & \textbf{0.13} & 100 & 0.31 & 100 & 0.35 & 100 \\ 
local\_unstable & -- & 37.1 & 53.98 & 100 & \textbf{0.40} & 100 & 0.62 & 100 \\  
varying\_robust & -- & 35.7 & 0.14 & 100 & 0.35 & 100 & \textbf{0.26} & 100 \\  
outdoor\_robust\_day1 & \textbf{1.43} & 100 & 2.92 & 100 & 2.55 & 100 & 2.15 & 100 \\ 
outdoor\_robust\_day2 & \textbf{2.18} & 100 & 2664.00 & 100 & 5.48 & 100 & 8.54 & 100 \\
outdoor\_robust\_night1 & \textbf{0.67} & 100 & 738.83 & 100 & 4.80 & 100 & 3.19 & 100 \\  
outdoor\_robust\_night2 & \textbf{0.50} & 100 & 605.62 & 100 & 3.17 & 100 & 6.53 & 100 \\
\hline
\end{tabular}
}
\end{table}

On the ViViD++ dataset~\cite{lee2022vivid++}, the proposed method demonstrates consistent performance across environments. ORB-SLAM3 achieves the lowest ATE in outdoor sequences but fails to initialize or maintain tracking in most indoor scenarios. The proposed method and DSO~\cite{engel2017direct} successfully complete the majority of sequences with comparable accuracy, though the proposed method yields lower ATE in more cases.  While ROTIO occasionally achieves good results, it often fails to recover the correct trajectory scale, limiting its overall reliability.

In summary, the proposed method demonstrates results that are in many cases better or comparable to those of the other evaluated approaches. It shows stable behavior in terms of initialization and tracking continuity, in contrast to ORB-SLAM3, which occasionally fails to initialize or maintain tracking. At the same time, ORB-SLAM3 benefits from the presence of a loop closure module, which improves its performance in certain scenarios, especially outdoors. Although ROTIO integrates IMU data, it does not consistently recover the correct trajectory scale. The DSO method generally maintains stable tracking and initialization but tends to yield lower accuracy compared to the other methods.

\subsection{Ablations on Weighted Factor-Graph}

This section analyzes the effect of incorporating keypoint confidence scores from SuperPoint into the factor graph optimization. The proposed weighted cost function is compared against a baseline using uniform weighting across all keypoints. The primary evaluation metric is the distribution of tracked frames before tracking loss, reported as median and interquartile range (IQR) percentages of the full trajectory length. As shown in Table~\ref{tab:weighting_results}, the confidence-weighted approach demonstrates improved tracking stability, maintaining trajectory estimation throughout all evaluated sequences. In contrast, the unweighted baseline exhibits premature tracking failure across multiple scenarios.

\begin{table}[]
\centering
\label{tab:weighting_results}
\caption{Median and IQR of tracked trajectory percentage for unweighted vs. weighted optimization}
\begin{tabular}{@{}lll@{}}
\toprule
Median/IQR (\%)               & ROTIO dataset & ViViD++  \\ \midrule
Uniform weighted FGO          & 93.1/17       & 100/75.7 \\
Confidence-based weighted FGO & 100/0         & 100/0    \\ \bottomrule
\end{tabular}
\end{table}

%% file: src/09_conclusion.tex
\section{Conclusion}

This paper presents a monocular graph-based SLAM system for thermal imagery that addresses key challenges namely scarcity of data for data-driven approaches and generality to the various thermal camera sensors. The proposed approach integrates a tailored preprocessing pipeline, general-purpose learned features (SuperPoint and LightGlue), and a confidence-weighted factor graph optimization that uses confidence scores from the neural detector. Experimental results demonstrate the key importance of the modifications to the visual SLAM pipeline and accordingly the system achieves robust performance across diverse thermal environments, outperforming existing methods in tracking continuity and accuracy while maintaining generalization without dataset-specific training. The work provides a practical solution for thermal SLAM in visually degraded conditions and establishes a foundation for future extensions involving multi-modal sensing by identifying key distinguishing characteristics between thermal and monocular color images.